\documentclass[letterpaper,10pt,journal,twoside]{IEEEtran}  

\usepackage{graphicx} 
\usepackage{placeins}
\usepackage{subcaption}
\usepackage{booktabs}
\usepackage{multirow}
\usepackage{amsmath} 
\usepackage{amssymb}  
\usepackage{dsfont}
\usepackage[most]{tcolorbox}
\usepackage{url}
\usepackage[hidelinks]{hyperref}

\tcolorboxenvironment{example}{
  enhanced,
  boxrule=0.6pt,
  colframe=black,
  colback=white,
  arc=1.5mm,
  left=1mm,right=1mm,top=0.8mm,bottom=0.8mm,
}

\newcommand{\Ind}{\mathds{1}} 

\newcommand{\bfs}[1]{
  \boldsymbol{#1}
}

\newcommand{\jp}{
  \bfs{q}
}
\newcommand{\jv}{
  \bfs{\dot{q}}
}
\newcommand{\ja}{
  \bfs{\ddot{q}}
}

\title{
\textcolor{blue}{SixthSense: Proprioception-Only Whole-Body Wrench Estimation for Humanoids}
}

\author{Xingzhou Chen$^{1,*}$, Xiayan Xu$^{1,*}$, Yan Ning$^{1}$, Jiyu Yu$^{2}$, Yizheng Zhang$^{3}$, Siyi Qian$^{3}$,\\
Lingzhu Xiang$^{3}$, Jiahao Chen$^{3}$, Ling Shi$^{1,\dagger}$, Yuquan Wang$^{3,\dagger}$, and Haodong Zhang$^{3,\dagger}$%
\thanks{$^{*}$Equal contribution. $^{\dagger}$Corresponding authors.}%
\thanks{$^{1}$Xingzhou Chen, Xiayan Xu, Yan Ning, and Ling Shi are with The Hong Kong University of Science and Technology, Hong Kong SAR, China.}%
\thanks{$^{2}$Jiyu Yu is with Zhejiang University, Hangzhou, China.}%
\thanks{$^{3}$Yizheng Zhang, Siyi Qian, Lingzhu Xiang, Jiahao Chen, Yuquan Wang, and Haodong Zhang are with Tencent Robotics X, Shenzhen, China.}%
}

\usepackage{xcolor}
\colorlet{blue}{black}

\AtBeginDocument{%
    \setlength{\abovedisplayskip}{4pt plus 1pt minus 1pt}%
    \setlength{\belowdisplayskip}{4pt plus 1pt minus 1pt}%
    \setlength{\abovedisplayshortskip}{1pt plus 1pt}%
    \setlength{\belowdisplayshortskip}{3pt plus 1pt minus 1pt}%
    \setlength{\jot}{3pt}%
    \setlength{\floatsep}{6pt plus 1pt minus 1pt}%
    \setlength{\textfloatsep}{8pt plus 1pt minus 1pt}%
    \setlength{\intextsep}{8pt plus 1pt minus 1pt}%
    \setlength{\dblfloatsep}{6pt plus 1pt minus 1pt}%
    \setlength{\dbltextfloatsep}{8pt plus 1pt minus 1pt}%
}
\patchcmd{\section}{1.5ex plus 1.5ex minus 0.5ex}%
    {0.8ex plus 0.5ex minus 0.2ex}{}%
    {\PackageWarning{compact-spacing}{Section spacing patch failed}}
\patchcmd{\section}{0.7ex plus 1ex minus 0ex}%
    {0.5ex plus 0.2ex minus 0ex}{}%
    {\PackageWarning{compact-spacing}{Section spacing patch failed}}
\patchcmd{\subsection}{1.5ex plus 1.5ex minus 0.5ex}%
    {0.8ex plus 0.5ex minus 0.2ex}{}%
    {\PackageWarning{compact-spacing}{Subsection spacing patch failed}}
\patchcmd{\subsection}{0.7ex plus .5ex minus 0ex}%
    {0.5ex plus 0.2ex minus 0ex}{}%
    {\PackageWarning{compact-spacing}{Subsection spacing patch failed}}
\AtBeginEnvironment{itemize}{%
    \setlength{\topsep}{1pt plus 0.5pt minus 0.5pt}%
    \setlength{\itemsep}{0pt}%
    \setlength{\parsep}{0pt}%
    \setlength{\partopsep}{0pt}%
}
\begin{document}

\maketitle

\begin{abstract}
Humanoid robots are entering the physical world at scale, yet are often criticized as oversized toys---good at dancing, but lacking the force-interaction capabilities for practical tasks. 
\begingroup\color{blue}
Reliable whole-body contact perception is difficult because floating-base
dynamics make proprioceptive wrench recovery ambiguous, while model-based
estimators require accurate dynamics and support-force inputs.
We propose SixthSense, which uses conditional flow matching to model sparse
spatiotemporal fields of link-wise contact gates and 6-D wrenches conditioned
on proprioceptive and IMU histories. We train one estimator on pooled data from
lying, standing, walking, and motion tracking.
Across extensive simulation experiments, SixthSense outperforms learned and
model-based baselines in overall contact estimation and remains robust to
observation noise. We further establish a real-robot benchmark with our newly
collected Unitree G1 dataset, where SixthSense achieves the strongest overall
performance among learned estimators and requires approximately 50\,ms per inference.
Controlled ambiguity experiments show that its samples
recover multiple dynamically plausible contact modes with improved local
physical consistency. We finally deploy SixthSense for collision detection and
contact-triggered physical human--robot interaction.
\par\endgroup
\end{abstract}

\section{Introduction}
\label{sec:intro}
Humanoid robots are increasingly capable of executing complex motions in the wild, from agile locomotion~\cite{hwangbo2019learning,lee2020learning} to expressive whole-body tracking~\cite{peng2018deepmimic,liao2025beyondmimic,ze2025twist}.
Yet physical contact---collisions, pushes and pulls, or carried loads---is treated as a disturbance to be rejected rather than a phenomenon to be perceived and understood.
As a result, robots remain unaware of when, where, and how much force is applied to their bodies, which limits their ability to engage in contact-rich interaction with the environment and humans~\cite{siciliano2008springer}.

This limitation is rooted in how current reinforcement-learning controllers handle external forces. Rather than estimating contact as a first-class quantity, they either absorb it into dense imitation objectives~\cite{peng2018deepmimic,liao2025beyondmimic,ze2025twist,he2024h2o}, compensate for it through history-conditioned policies~\cite{kumar2021rma,zhang2025track}, or expose it only through torque-feedback manipulation pipelines~\cite{zhang2025elucidating}.
\textcolor{blue}{Thus, contact remains embedded in task-specific control rather than exposed for downstream use.}

Explicit external-wrench estimation is well established for fixed-base manipulators through residual-based observers~\cite{de2005sensorless}. Humanoid whole-body contact, however, violates the underlying assumptions: a floating base and multiple contact modes render the problem under-determined and nearly unobservable~\cite{de2006collision,haddadin2008collision,flacco2016residual,haddadin2017robot}. Existing model-based observers and filters thus hinge on idealized dynamics, well-conditioned Jacobians, and precise ankle force measurements that rarely hold on real humanoids.
Dense force or tactile sensing on whole-body humanoids is limited by hardware cost, calibration complexity, and limited body coverage~\cite{yuan2017gelsight,yan2021soft,lambeta2020digit}. \textcolor{blue}{Learning-based estimators instead focus on end-effector forces or robot-arm contacts rather than humanoid whole-body contact.}
\textcolor{blue}{Contact is also ambiguous: distinct whole-body contacts can induce nearly identical proprioceptive responses, making contact recovery from motion non-identifiable. Consequently, shared estimation of contact timing, location, and wrench across humanoid behaviors remains underexplored.}

\begin{figure}
    \centering
    \includegraphics[width=1.0\linewidth]{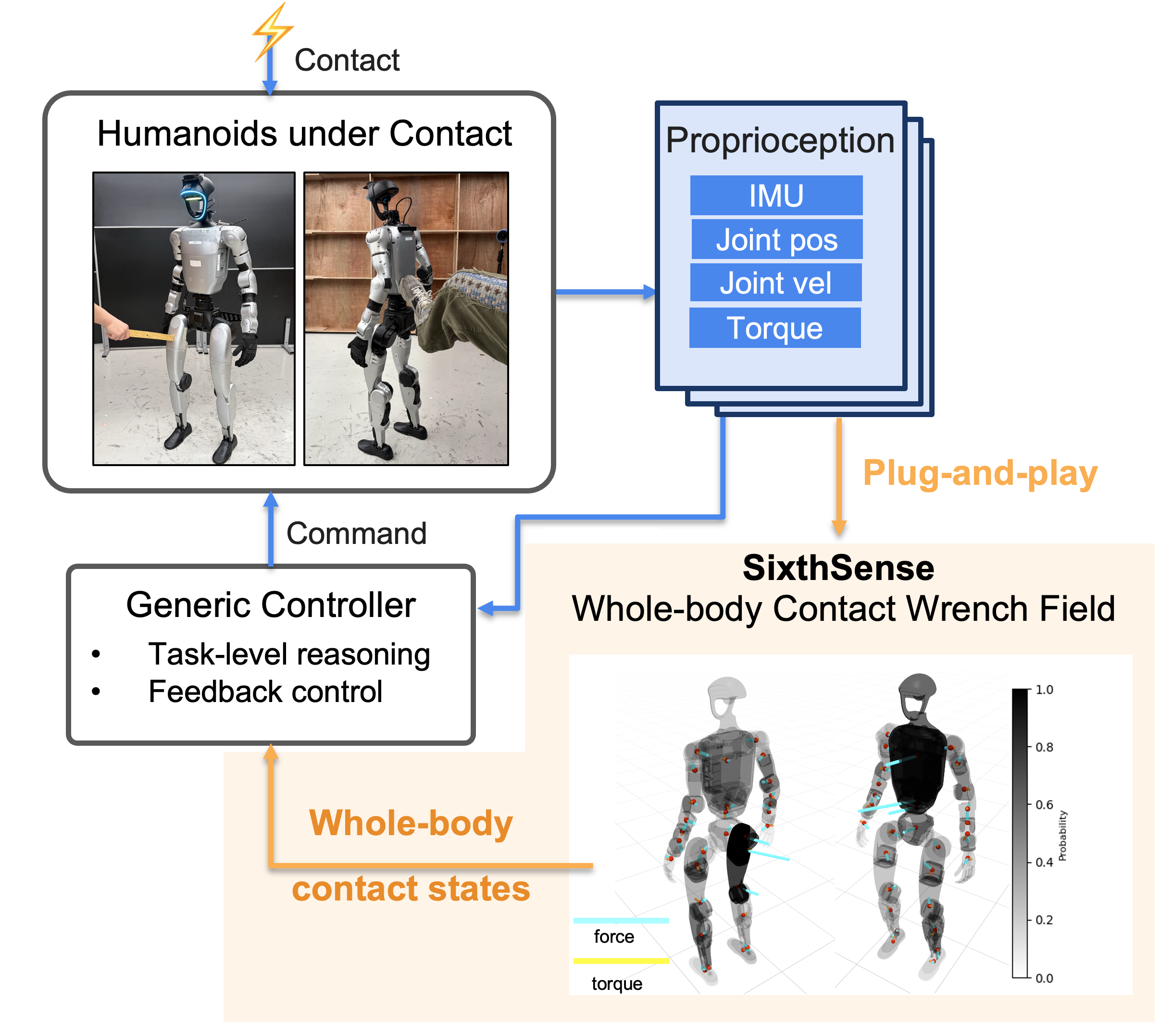}
    \caption{{\bf SixthSense}: Inferring whole-body contact wrench fields via proprioception. \textcolor{blue}{A shared estimator provides contact estimates across the evaluated controller regimes through a common input/output interface.}}
    \label{fig:overview}
\end{figure}

We cast whole-body contact estimation as conditional generative inference~\cite{kingma2013auto,song2020score} over a spatiotemporal distribution of external wrenches conditioned on proprioception alone. Inspired by human sensorimotor experience, we term this form of bodily awareness \textbf{SixthSense}, a perceptual modality complementary to vision or touch sensing~\cite{calandra2018more,suresh2024neuralfeels,lee2020making}. As illustrated in Fig.~\ref{fig:overview}, the external contact wrench is represented as a sparse spatiotemporal field that jointly encodes contact likelihood and wrench value, and its conditional distribution is learned via conditional flow matching (CFM)~\cite{lipman2022flow} from temporally aligned proprioception. The field representation unifies detection and estimation in a single output, while the learned CFM mapping captures the multi-modal posterior that arises from the inherent ambiguity of whole-body dynamics, \textcolor{blue}{providing a contact-perception module} that answers \textit{whether}, \textit{where}, \textit{when}, and \textit{how much} non-foot contact occurs.

\textcolor{blue}{We evaluate the shared estimator across lying, standing,
walking, and whole-body tracking controllers. In simulation, we achieve a
better balance of detection, localization, and wrench accuracy than model-based
and learned baselines, while remaining robust to observation noise. Controlled
ambiguity tests show broader coverage of plausible contact modes and more
consistent local replays than a generative baseline. On our collected Unitree
G1 benchmark, we obtain fewer false positives and better localization,
temporal coverage, and wrench estimates than learned baselines. With
approximately 50\,ms inference, we also demonstrate collision detection and
contact-triggered pHRI.}

Our contributions are threefold:
\begin{itemize}
    \item \textcolor{blue}{We formulate humanoid whole-body contact estimation
    as generative inference over a sparse spatiotemporal wrench field
    conditioned on proprioception.}
    \item \textcolor{blue}{We develop a shared CFM estimator that jointly
    outputs link-wise contact gates and 6-D wrenches, samples distinct contact
    locations and wrenches under proprioceptive ambiguity, and uses a common
    interface across multiple controllers.}
    \item \textcolor{blue}{We establish a Unitree G1 contact-estimation
    benchmark and validate SixthSense across simulation and real-robot tests,
    including noise and ambiguity evaluations and contact-triggered pHRI.}
\end{itemize}

\section{Related Work}
\label{sec:related_work}

\subsection{Implicit Contact-Aware Humanoid Motion Control}
Classical humanoid pipelines treat accurate and timely contact information as essential for balancing locomotion and multi-contact manipulation~\cite{caron2019tro,hereid2019iros,gong2019acc}, with extensions to uneven terrain~\cite{mesesan2019humanoid,gong2019acc} and contact-rich industrial settings~\cite{kheddar2019ram}.
Recent reinforcement-learning controllers absorb contact handling into the policy itself, embedding floating-base state estimation in a learned network and yielding stable foothold behaviors and disturbance recovery from IMU and joint feedback alone~\cite{ji2022ral}.
For motion tracking, imitation-based approaches induce physically consistent contacts and recovery behaviors through dense tracking objectives, while more recent systems scale interaction diversity via diffusion-based motion synthesis or large-scale whole-body teleoperation~\cite{peng2018deepmimic,liao2025beyondmimic,ze2025twist,he2024h2o}.
In parallel, disturbance-aware tracking methods treat external interactions as latent disturbances to be compensated online, using history-conditioned representations for adaptation~\cite{kumar2021rma,zhang2025track}.
Related ideas also appear in end-effector manipulation, where joint torque feedback is incorporated into Vision-Language-Action models to improve performance in contact-rich tasks without dedicated force sensing~\cite{zhang2025elucidating}.
\textcolor{blue}{In these approaches, contact handling is incorporated into control, rather than exposed through a separate whole-body contact-estimation interface for downstream modules.}
\subsection{Sensor-Based Contact-Rich Control}
Direct force and tactile sensing has been extensively studied in dexterous manipulation, where high-resolution tactile and force sensors enable accurate estimation of contact location, geometry, and force, supporting grasp stability and fine interaction control~\cite{yuan2017gelsight,yan2021soft,lambeta2020digit}.
In a complementary direction, visuotactile perception fuses vision with touch for in-hand reasoning~\cite{calandra2018more,suresh2024neuralfeels,lee2020making}, and humanoid pipelines exploit visual cues to anticipate task-space contact geometry~\cite{lee2020learning,ze2025twist2}.
However, dense force and tactile sensing remains impractical to scale to whole-body humanoids due to high hardware cost, and visual cues only target task-space interaction objects rather than body-level force interaction during execution.
\subsection{Sensorless Contact Force Estimation}
Dynamics-residual methods estimate external wrenches on fixed-base
manipulators and floating-base robots~\cite{de2005sensorless,de2006collision,haddadin2008collision,flacco2016residual,haddadin2017robot,manuelli2016cpf}.
The generalized momentum observer (GMO) estimates external generalized
forces without acceleration measurements~\cite{haddadin2017robot},
while the contact particle filter (CPF) tracks multiple contact
hypotheses through particle filtering~\cite{manuelli2016cpf}.
These methods require accurate dynamics, well-conditioned Jacobians,
and precise ankle-force measurements that rarely hold on real humanoids.
CPF also requires a prescribed contact count.
Related foot-contact observers~\cite{payne2024multi,hwangbo2016iros,camurri2017probabilistic}
and legged state estimators~\cite{bloesch2013state,hartley2020contact}
target locomotion or base-state estimation rather than whole-body wrenches.

\begingroup\color{blue}
Learning-based methods estimate end-effector forces for
loco-manipulation~\cite{zhi2025learning}.
NeuralActuator~\cite{dou2026neuralactuator} predicts arm actuation and
external forces from command/state histories and actuator telemetry.
CDM~\cite{han2025cdm} samples contact locations using diffusion.
ContactFlowFilter~\cite{han2026mixture} combines flow matching and
particle filtering to track contact count, locations, and 3D forces.
Both address multimodal arm contacts using joint torque and base
force/torque sensing.
We instead model a joint temporal distribution of link-wise contact
gates and 6-D wrenches for whole-body non-foot contacts on floating-base
humanoids, using proprioceptive and IMU histories without policy actions
or base force/torque sensing.
\par\endgroup

\section{Problem Statement}
\label{sec:problem}
\subsection{Dynamics of Humanoid Robots}
A floating-base robot has generalized coordinates $\jp=[\jp_{\mathrm{b}}^\top\;\jp_{\mathrm{j}}^\top]^\top$, where $\jp_{\mathrm{b}}\in\mathbb{R}^6$ locally parameterizes the base pose and $\jp_{\mathrm{j}}\in\mathbb{R}^n$ contains $n$ joint angles. Neglecting joint friction, the rigid-body dynamics are given by
\begin{equation}
\label{eq:dyn}
H(\jp)\ja + C(\jp,\jv)\jv + g(\jp)
=  S^\top\boldsymbol{\tau}_{\mathrm{m}} + \sum_{e=1}^{N_c} J_e(\jp)^\top \boldsymbol{f}_e ,
\end{equation}
where $H$ is the inertia matrix and $C\jv+g$ collects Coriolis, centrifugal, and gravity terms. The selection matrix $S=[\,\boldsymbol{0}_{n\times 6}\;\mathbf{I}_n\,]$ maps joint torques $\boldsymbol{\tau}_{\mathrm{m}}$ to generalized forces.
Each contact wrench $\boldsymbol{f}_e\in\mathbb{R}^6$ contributes to generalized forces linearly through its Jacobian $J_e(\jp)$.

Adopting the base--joint block form~\cite{featherstone2008rigid}, we have
\begin{align}
\label{eq:dyn_block}
\begin{bmatrix}
    H_{\mathrm{bb}} & H_{\mathrm{bj}}\\
    H_{\mathrm{bj}}^\top & H_{\mathrm{jj}}
\end{bmatrix}
\begin{bmatrix}
    \ja_{\mathrm{b}}\\
    \ja_{\mathrm{j}}
\end{bmatrix}
+
\begin{bmatrix}
    \boldsymbol{h}_{\mathrm{b}}\\
    \boldsymbol{h}_{\mathrm{j}}
\end{bmatrix}
- S^\top\boldsymbol{\tau}_{\mathrm{m}}
=
\sum_{e=1}^{N_c}
\begin{bmatrix}
    J_{e,\mathrm{b}}^\top\\
    J_{e,\mathrm{j}}^\top
\end{bmatrix}
\boldsymbol{f}_e,
\end{align}
where the subscripts $\mathrm{b}$ and $\mathrm{j}$ index the base and joint blocks of $H(\jp)$ and of $\boldsymbol{h}=C(\jp,\jv)\jv+g(\jp)$.

\subsection{Problem of Interest}
\begingroup\color{blue}
Let $\mathbf{o}_t$ denote the proprioceptive observation at time $t$, and $\mathcal{O}_{t:t+T-1}=\{\mathbf{o}_k\}_{k=t}^{t+T-1}$ a window of length $T$, with
\begin{equation}
    \label{eq:observ}
    \mathbf{o}_t
    =
    \left[ \mathbf{q}_t,\, \dot{\mathbf{q}}_t,\, \boldsymbol{\tau}_t,\, \boldsymbol{\omega}_{\mathrm{b},t},\, \mathbf{g}_{\mathrm{b},t},\, \mathbf{a}_{\mathrm{b},t}
    \right]
    \in\mathbb{R}^{96},
\end{equation}
where $\mathbf{q}_t$, $\dot{\mathbf{q}}_t$, and $\boldsymbol{\tau}_t$ denote the joint positions, velocities, and motor-torque estimates, respectively, while $\boldsymbol{\omega}_{\mathrm{b},t}$, $\mathbf{g}_{\mathrm{b},t}$, and $\mathbf{a}_{\mathrm{b},t}$ denote the IMU angular velocity, projected gravity, and gravity-compensated linear acceleration.
In~\eqref{eq:dyn_block}, the left-hand side is fully determined by $\mathcal{O}_{t:t+T-1}$
and a known rigid-body model, whereas the right-hand side depends on a latent contact configuration through the unknown external wrenches.

Analytical approaches~\cite{haddadin2008collision,flacco2016residual,haddadin2017robot} first localize contacts to construct $J_e(\jp)$, then estimate $\boldsymbol{f}_e$ through inverse dynamics or residual formulations.
Under contact uncertainty, the mapping from wrenches to joint- and base-level residuals becomes non-injective, admitting nullspace solutions and rendering wrench recovery structurally non-identifiable~\cite{flacco2016residual}.
Modeling errors and sensor noise are further amplified by ill-conditioned Jacobians, limiting analytical estimation in practice.
We therefore lift contacts to $N$ candidate links rather than selecting an explicit contact set, approximating the generalized contact wrench as
\begin{equation}
\label{eq:dyn_lift}
\sum_{e=1}^{N_c} J_e^\top \boldsymbol{f}_e
\;\approx\;
\mathbf{J}_{\mathrm{ext}}^\top\,\mathbf{f}_{\mathrm{ext}},
\end{equation}
where
$\mathbf{J}_{\mathrm{ext}}^\top=[(J^{(1)})^\top,\ldots,(J^{(N)})^\top]\in\mathbb{R}^{(6+n)\times 6N}$
and
$\mathbf{f}_{\mathrm{ext}}\in\mathbb{R}^{6N}$ stacks the corresponding candidate wrenches.
\par\endgroup

This lifting raises three questions for whole-body wrench perception:
\begin{itemize}
    \item \textit{Representation.} How should sparse, multi-point whole-body wrenches be represented so that whether, where, and how much are jointly captured over space and time?
    \item \textit{Inference.} How can wrenches be inferred when distinct contacts produce nearly identical proprioceptive responses?
    \item \textcolor{blue}{\textbf{Multi-controller estimation.} As dynamics vary with tasks, gaits, and support configurations, how can one estimator handle multiple behaviors and controller regimes represented in its training data?}
\end{itemize}
In this work, we address these questions through conditional generative inference over a spatiotemporal wrench field.

\begin{figure*}[htp]
    \centering
    \includegraphics[trim=0 28bp 0 14bp,clip,width=1.00\linewidth]{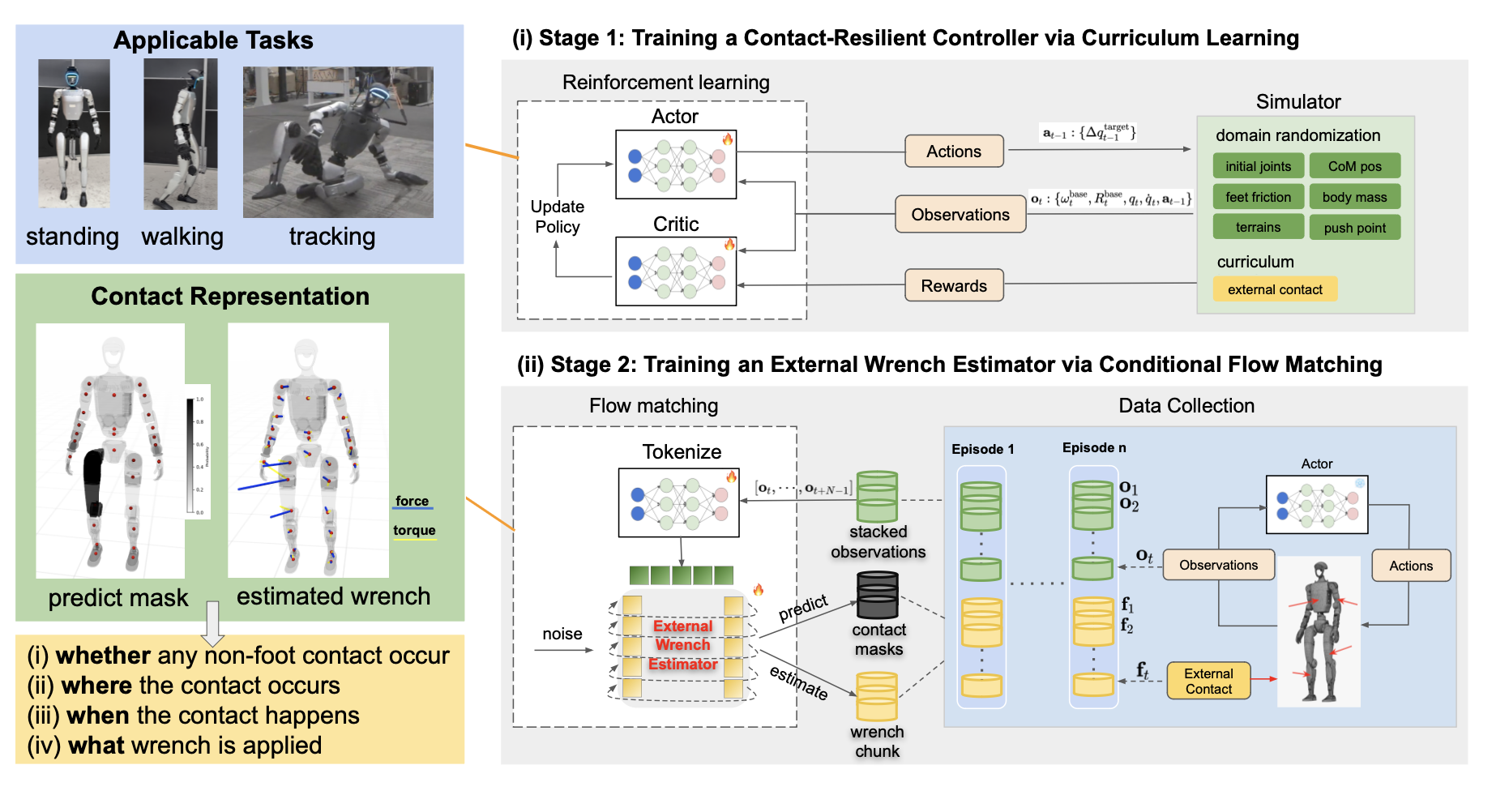}
    \caption{Overview: 
    Given a contact-resilient control policy, we use its rollouts to train a conditional flow-matching model that outputs a whole-body external contact wrench field over the discretized robot surface.
    }
    \label{fig:contact_pipeline}
\end{figure*}

\section{Method}
\label{sec:method}


\subsection{Contact Representation}
\label{subsec:represent}

\begingroup\color{blue}
Existing analytical contact-estimation methods~\cite{haddadin2017robot}
commonly separate contact detection, localization, and force estimation.
Early hard decisions can propagate errors to later stages, especially under modeling inaccuracies or simultaneous contacts.
\par\endgroup

\begingroup\color{blue}
We instead represent the robot body by $N$ candidate links.
At each frame $t$, the model predicts continuous gate scores $\widehat{\mathbf{G}}_t\in\mathbb{R}^{N}$ and a wrench field $\widehat{\mathbf{W}}_t\in\mathbb{R}^{N\times6}$.
For link $i$,
\begin{equation}
    \widehat{\mathbf{W}}_t^{(i)}
    =
    \begin{bmatrix}
        \widehat{\mathbf{f}}_t^{(i)}\\
        \widehat{\boldsymbol{\tau}}_t^{(i)}
    \end{bmatrix}.
\end{equation}
Each wrench is expressed about the corresponding link center of mass, with its components resolved in the pelvis frame.
This representation permits several links to be active simultaneously.
\par\endgroup

\begingroup\color{blue}
The gated wrench output is
\begin{equation}
    \label{eq:decision}
    \widetilde{\mathbf{W}}_t^{(i)}
    =
    \Ind\!\left[
    \widehat{\mathbf{G}}_t^{(i)}\geq\delta
    \right]
    \Ind\!\left[
    \left\|\widehat{\mathbf{f}}_t^{(i)}\right\|_2
    \geq f_{\min}
    \right]
    \widehat{\mathbf{W}}_t^{(i)},
\end{equation}
where $\Ind[\cdot]$ is the indicator function, $\delta$ is the gate threshold, and $f_{\min}$ is the minimum force-magnitude threshold.

\par\endgroup

\subsection{Proprioceptive Conditioning and Observability Hypotheses}
\label{subsec:proprio}

\begingroup\color{blue}
Given the proprioceptive window
$\mathcal{O}_{t:t+T-1}$, whose observations are defined
in~\eqref{eq:observ}, a learnable temporal encoder $g_\phi$
produces the conditioning sequence
\begin{equation}
    \mathbf{c}
    =
    g_\phi(\mathcal{O}_{t:t+T-1}) \in\mathbb{R}^{T\times d_c},
\end{equation}
where $\phi$ denotes the encoder parameters and $d_c$ is the feature dimension.
We use $T=50$ frames sampled at $50\,\mathrm{Hz}$ and $d_c=160$.
The resulting sequence conditions the joint gate--wrench flow introduced in Sec.~\ref{sec:cfm}.
\par\endgroup

\begingroup\color{blue}
External contacts perturb the robot's motion and joint dynamics, leaving signatures in its proprioceptive history.
This motivates two working hypotheses:
\par\endgroup

\begingroup\color{blue}
\noindent\textit{\textbf{Hypothesis 1 (Observability).}}
\textit{Contacts that produce a measurable closed-loop response are at least partially observable from the proprioceptive history.}
\par\endgroup

\begingroup\color{blue}
\noindent\textit{\textbf{Hypothesis 2 (Robustness--Observability Coupling).}}
\textit{A contact-resilient controller produces more structured and repeatable responses to external disturbances, making contact-related features easier to decode.}
\par\endgroup

\begingroup\color{blue}
Based on these hypotheses, we adopt a two-stage procedure.
In Stage~1, we train a contact-resilient control policy using domain randomization and a curriculum over diverse external disturbances and contact configurations.
In Stage~2, we train the joint gate--wrench estimator conditioned on $\mathbf{c}$ (Sec.~\ref{sec:cfm}).
\par\endgroup

\subsection{Joint Contact Inference via Conditional Flow Matching}
\label{sec:cfm}

\begingroup\color{blue}
Different contact configurations can produce similar proprioceptive responses.
We therefore use conditional flow matching (CFM)~\cite{tong2023improving} to model the joint distribution of contact gates and wrenches conditioned on the encoded proprioceptive sequence $\mathbf{c}$.
At each frame $t$, the target concatenates the gate targets $\mathbf{G}_t$ and normalized per-link wrenches:
\begin{align}
    \mathbf{y}_t
    &= \operatorname{col}\!\left(
    \mathbf{G}_t,
    \overline{\mathbf{W}}_t^{(1)},\ldots,
    \overline{\mathbf{W}}_t^{(N)}
    \right), \\
    \overline{\mathbf{W}}_t^{(i)}
    &=
    \begin{bmatrix}
        \mathbf{f}_t^{(i)}/s_f \\
        \boldsymbol{\tau}_t^{(i)}/s_\tau
    \end{bmatrix},
\end{align}
where $\operatorname{col}$ denotes vertical concatenation and $s_f,s_\tau>0$ are force and torque scales.
Thus $\mathbf{y}_t\in\mathbb{R}^{7N}$, and stacking the frame targets gives $\mathbf{y}\in\mathbb{R}^{T\times7N}$.
\par\endgroup

\begingroup\color{blue}
Given the joint target $\mathbf{y}$, we independently draw $\mathbf{z}$ and $\boldsymbol{\epsilon}$ from $\mathcal{N}(\mathbf{0},\mathbf{I})$ and define the training path
\begin{equation}
\label{eq:cfm-path}
    \mathbf{y}^{\sigma}
    =
    \mathbf{y}+\sigma_{\mathrm{end}}\boldsymbol{\epsilon},
    \qquad
    \mathbf{x}_s
    =
    (1-s)\mathbf{z}+s\mathbf{y}^{\sigma},
\end{equation}
where $s\in[0,1)$ is the sampled flow time and $\sigma_{\mathrm{end}}$ controls the amount of target smoothing.
The intermediate state $\mathbf{x}_s$ interpolates between Gaussian noise and the perturbed target.
\par\endgroup

\begingroup\color{blue}
\paragraph{Network architecture}
At each frame $t$, the network combines the conditioning feature
$\mathbf{c}_t$ and flow state $\mathbf{x}_{s,t}$ into a token:
\begin{equation}
    \mathbf{h}_t
    =P_c\!\left(\operatorname{LN}(\mathbf{c}_t)\right)
    +P_x(\mathbf{x}_{s,t})
    +\mathbf{e}(s)
    +\mathbf{p}_t
    \in\mathbb{R}^{192},
\end{equation}
where $P_c$ and $P_x$ are learned linear projections, $\operatorname{LN}$ is layer normalization, $\mathbf{e}(s)$ embeds Fourier-encoded flow time through an MLP, and $\mathbf{p}_t$ encodes temporal position.
Four temporal self-attention blocks, each with four heads, process these tokens; layer normalization and a linear projection then produce the joint endpoint prediction:
\begin{equation}
\label{eq:endpoint}
\widehat{\mathbf{y}}
= D_\theta(\mathbf{x}_s,s;\mathbf{c})
\in\mathbb{R}^{T\times7N}.
\end{equation}
\par\endgroup

\begingroup\color{blue}
\paragraph{Training}
The network predicts the endpoint in~\eqref{eq:endpoint} and minimizes
\begin{equation}
\label{eq:loss}
    \mathcal{L}_{\mathrm{CFM}}
    =
    \mathbb{E}\left[
    \operatorname{MSE}
    (\widehat{\mathbf{y}}_G,\mathbf{y}^{\sigma}_G)
    +
    \lambda_W
    \operatorname{MSE}
    (\widehat{\mathbf{y}}_W,\mathbf{y}^{\sigma}_W)
    \right],
\end{equation}
where $G$ and $W$ denote gate and normalized wrench
coordinates, respectively.
Each MSE averages over valid supervised entries, and $\lambda_W>0$ balances the gate and wrench terms.
\par\endgroup

\begingroup\color{blue}
The endpoint prediction defines the velocity field
\begin{equation}
\label{eq:cfm-velocity}
    \mathbf{v}_\theta(\mathbf{x}_s,s;\mathbf{c})
    =
    \frac{
    D_\theta(\mathbf{x}_s,s;\mathbf{c})-\mathbf{x}_s
    }{1-s}.
\end{equation}
Under the linear path in~\eqref{eq:cfm-path}, the endpoint loss is equivalent to a velocity-matching loss weighted by $(1-s)^2$; training therefore requires no numerical integration.
\par\endgroup

\begingroup\color{blue}
\paragraph{Inference}
Given $\mathbf{c}$, we draw $\mathbf{x}_0\sim\mathcal{N}(\mathbf{0},\mathbf{I})$ and integrate the velocity field in~\eqref{eq:cfm-velocity}
from $s=0$ to $s=1$ using 24 steps of the explicit midpoint (second-order Runge--Kutta) method.
The terminal state $\mathbf{x}_1$ provides gate scores and normalized wrenches.
We rescale the wrench components by $s_f$ and $s_\tau$ and apply~\eqref{eq:decision} to obtain the reported contact wrenches.
Each window uses one independent joint sample.
\par\endgroup

\begingroup\color{blue}
Inference runs every five sensor frames (10 Hz) over the latest
50-frame window (1 s).
We report the center-frame wrench estimates with a nominal
0.5 s delay and use the final-frame prediction for contact detection.
Predictions from overlapping windows are smoothed.
\par\endgroup

\section{Experiment}
\begingroup\color{blue}
We evaluate single-contact estimation on simulated test data and Unitree G1 recordings, then examine whether repeated samples capture distinct contact hypotheses under controlled ambiguity. We further illustrate contact-triggered physical human--robot interactions. The baselines and metrics are specified in the corresponding subsections.
\par\endgroup

\begingroup\color{blue}
\subsection{Contact Estimation in Simulation}
\label{subsec:expa}

\textbf{Data.}
We collect 100,000 single-contact episodes in simulation across different
tasks, with 20\% each for standing, walking, and lying, and 40\% for motion tracking.
Each episode contains one contact event, preceded and followed by 3\,s
of contact-free motion.
Contact links are sampled among 30 links in proportion to mesh surface
area, with uniformly random force directions.
We uniformly sample pulse wrench magnitudes from 12--80\,N for force
and 0.1--8\,N$\cdot$m for torque, and durations from 0.4--1.6\,s,
then apply wrenches as trapezoidal pulses at the link's center of mass.

We split the episodes into 90\% for training and 10\% for testing.
During training, we add zero-mean Gaussian observation noise with
 standard deviations of $0.5^\circ$ (orientation), $0.01$\,rad/s
(angular velocity), $0.1$\,m/s$^2$ (acceleration), $0.005$\,rad (joint
position), $0.05$\,rad/s (joint velocity), and $0.5$\,N$\cdot$m (joint
torque).
During testing, we multiply these standard deviations by a scale factor
$\alpha$ to simulate varying noise levels,
including extreme conditions.

We train a single SixthSense estimator on pooled rollouts from lying in damping mode, standing, walking, and TWIST2 motion tracking. The same trained weights are used across all four simulation tasks, rather than training a separate estimator for each controller. The episode-level split evaluates performance within these controller regimes rather than unseen controllers.


\textbf{Baselines.}
The learning baselines are \emph{SixthSense w/o CFM}, which retains our
backbone but directly regresses contact masks and wrenches, and a
conditional variational autoencoder (CVAE)~\cite{sohn2015conditional}.
We draw CVAE predictions from its conditional prior.
All receive the same 50-frame proprioceptive windows.
We also report GMO~\cite{haddadin2017robot} and
CPF~\cite{manuelli2016cpf}, supplied with simulated rigid-body state,
actuation, and ground-support inputs with foot-wrench noise of 10\,N and 1\,N$\cdot$m at $\alpha=1$. 
These model-based methods serve as references with additional information.
SixthSense, CVAE, and CPF use 3 independent draws; w/o CFM and GMO are deterministic.

\textbf{Metrics.}
Detection rate (DR) is the fraction of true-contact records with at least
one predicted contact during the valid contact interval, averaging draws
within each record before averaging records.
False positive rate (FPR) is the pooled fraction of quiet frames with a
predicted contact. Exact-link localization accuracy (LA) is the pooled
fraction of detected true-contact frames with exact link agreement.
Correctly localized temporal coverage ($R_{\mathrm{cov}}$) divides exact
localization hits by all true-contact frames, including misses.
Contact-duration ratio ($R_l$) compares predicted and true contact duration
on the true target link; 1 is ideal.
Force and moment mean-magnitude absolute errors ($e_F$ and $e_M$) compare
predicted and true mean norms over their respective nonzero valid intervals.
Duration and mean-magnitude error metrics are averaged over defined records
after averaging draws. Wrenches use pelvis axes and moments about each
link origin.

\textbf{Results.}
As shown in Table~\ref{tab:expa}, at the training noise level ($\alpha=1$),
SixthSense provides the strongest overall contact estimation performance
among the proprioception-only methods.
Among the proprioception-only methods, SixthSense has the highest DR
(91.94\%, versus 77.63\% and 89.49\%), the lowest pooled FPR
(0.15\%, versus 1.02\% and 3.41\%), the highest exact-link LA
(34.96\%, versus 30.25\% and 31.68\%), and the highest $R_{\mathrm{cov}}$
(25.48\%, versus 16.23\% and 21.25\%).
Its detection rate approaches that of GMO and CPF, which require
additional inputs. Its mean-magnitude errors are 9.42\,N and
2.57\,N$\cdot$m: the lowest force error among all methods, while
CVAE has the lowest moment error (2.37\,N$\cdot$m).

\begin{table}[ht]
\centering
\begingroup
\small
\setlength{\tabcolsep}{2pt}
\renewcommand{\arraystretch}{1.2}
\begin{tabular*}{\columnwidth}{@{\extracolsep{\fill}}lccccc@{}}
\toprule
& \multicolumn{3}{c}{Proprioception only} & \multicolumn{2}{c}{Privileged inputs} \\
\cmidrule(lr){2-4}\cmidrule(l){5-6}
Metric & SixthSense & w/o CFM & CVAE & GMO & CPF \\
\midrule
DR $\uparrow$ (\%) & \textbf{91.94} & 77.63 & 89.49 & 93.24 & 93.24 \\
FPR $\downarrow$ (\%) & \textbf{0.15} & 1.02 & 3.41 & 0.89 & 0.89 \\
LA $\uparrow$ (\%) & \textbf{34.96} & 30.25 & 31.68 & 19.32 & 21.16 \\
$R_{\mathrm{cov}}\uparrow$ (\%) & \textbf{25.48} & 16.23 & 21.25 & 8.80 & 9.63 \\
$R_l$ ($\to 1$) & \textbf{0.25} & 0.18 & 0.24 & 0.09 & 0.10 \\
$e_F\downarrow$ (N) & \textbf{9.42} & 14.90 & 10.89 & 9.53 & 9.54 \\
$e_M\downarrow$ (N$\cdot$m) & 2.57 & 2.54 & \textbf{2.37} & 8.96 & 9.23 \\
\bottomrule
\end{tabular*}
\endgroup

\caption{Simulation contact-estimation results. Bold marks the best
proprioception-only value in each row. GMO and CPF are excluded from this
comparison because they use privileged inputs.}
\label{tab:expa}
\end{table}


\textbf{Robustness to observation noise.}
On the same test records, we scale the training observation noise by
$\alpha\in\{0,0.25,0.5,1,1.5,2,5\}$, sharing each noise realization
across methods. GMO and CPF also receive noisy foot
wrenches ($2$\,N and $0.2$\,N$\cdot$m at $\alpha=1$).
Models remain fixed, while detection thresholds are adjusted adaptively
to balance DR and FPR. We use a tolerant
$\mathrm{LA}_{\leq1}$ criterion, allowing one kinematic-tree edge of
localization error. Fig.~\ref{fig:expa-observation-noise} shows that SixthSense's
$\mathrm{LA}_{\leq1}$ decreases by only $4.1\%$ from
$\alpha=0$ to $\alpha=2$, while its FPR remains below $1\%$. Even under severe observation noise ($\alpha=5$), SixthSense achieves
the most robust performance among all evaluated methods.

\begin{figure}[ht]
    \centering
    \includegraphics[width=\linewidth]{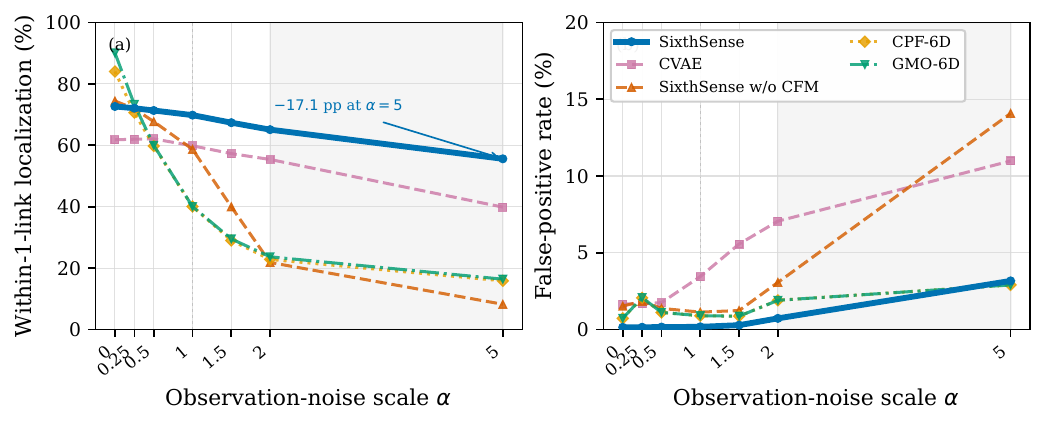}
    \caption{Observation-noise robustness on simulation evaluation. (a) Tolerant localization and (b) false-positive rate. $\mathrm{LA}_{\leq1}$ is conditioned on detection; the dotted line marks the base noise level ($\alpha=1$), and the shaded region is the additional stress range beyond the main sweep.}
    \label{fig:expa-observation-noise}
\end{figure}


\endgroup

\begingroup\color{blue}
\subsection{Contact Estimation on Real-Robot}
\label{subsec:expb}

\begin{figure}[t]
\centering
\includegraphics[width=0.8\linewidth]{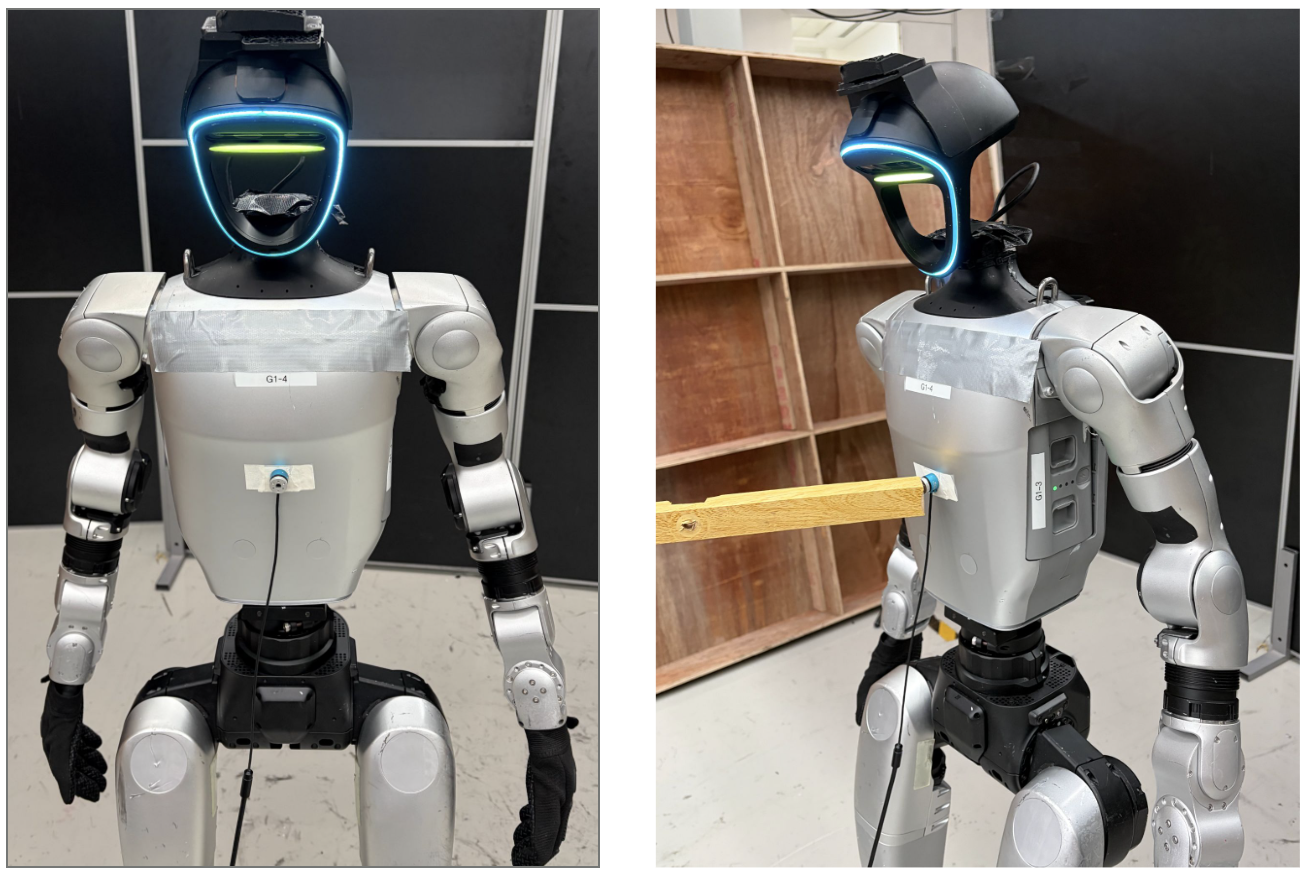}
\caption{Real-robot contact data collection using a 6-axis force/torque sensor.}
\label{fig:expb}
\end{figure}

\textbf{Hardware setup.}
We select 14 representative contact points on a Unitree G1 robot,
distributed across 12 distinct links, including torso, shoulder, elbow,
and hip links.
To measure contact wrenches, we use a single SRI M3701C 6-axis force/torque
sensor with rated capacities of 200\,N for $F_x$ and $F_y$,
400\,N for $F_z$, and 2\,N$\cdot$m for each torque axis,
with 1\% full-scale nonlinearity.
We remount the same sensor at different contact points on the robot's
surface and mark each mounting location.
As shown in Fig.~\ref{fig:expb}, we use a rod to apply contact forces
to the robot through the sensor,
pressing approximately normal to the sensor surface to reduce slippage.
The sensor is connected by cable to a Mac laptop, which records its
measurements at 100\,Hz. We synchronize the clocks of the laptop and
the robot's onboard computer and align the sensor measurements with
the proprioceptive ROS bag data using their timestamps.
In this deployment, model inference takes approximately 50\,ms with
the 24 CFM integration steps specified in Sec.~\ref{sec:cfm}.
This is computation time, separate from the observation-window delay.
The step count can be reduced for tasks requiring faster inference,
with estimation accuracy checked under the chosen setting.
\par\smallskip

\textbf{Data.}
We collect 372 recordings\footnote{Our real-robot dataset is available at
\url{https://huggingface.co/datasets/BlablaO/SixthSense-G1-Contact-Dataset}.}
on a Unitree G1 across three scenarios:
312 while standing, 30 while walking, and 30 while lying.
In the standing trials, we apply brief contacts at the sensor
location. In the walking trials, the moving robot collides with a mannequin. In the
lying trials, we pull a single link without instrumented wrench ground truth.
We record proprioception in ROS bag files and capture video for all trials,
with force-sensor output available for standing trials. After processing,
each recording
contains 3--6\,s of proprioceptive data sampled at 50\,Hz.
For standing, we also record the contact position and corresponding link, and convert the measured surface contact force into an equivalent wrench at the link's center of mass.
For walking and lying, contact intervals are identified from video
and motion annotations; no force or moment measurements are available.

\textbf{Results.}
We fine-tune SixthSense, w/o CFM, and CVAE, initially trained solely on simulation,
using 24 real-robot standing recordings (2 per link), and evaluate them on the remaining 288 standing recordings.
The CVAE gate threshold was selected on these evaluation recordings.
Table~\ref{tab:expbb} shows that SixthSense provides more reliable
contact estimation than the baselines on the real robot.
SixthSense achieves the lowest FPR$^{*}$ (0.66\%), the highest
exact-link LA (66.89\%), and the highest $R_{\mathrm{cov}}$ (57.14\%).
Its force and moment mean-magnitude errors are also the lowest
(4.61\,N and 0.63\,N$\cdot$m).
Although w/o CFM achieves a higher DR than SixthSense
(91.04\% versus 86.70\%), it incurs a much higher FPR$^{*}$
(8.05\% versus 0.66\%).
For SixthSense, the higher localization accuracy and temporal coverage
and lower force error in Table~\ref{tab:expbb} than in
Table~\ref{tab:expa} may result from forces being applied nearly
perpendicular to the surfaces, making contacts easier to distinguish.


\begin{table}[t]
\centering
\begingroup
\small
\setlength{\tabcolsep}{4pt}
\renewcommand{\arraystretch}{1.2}
\begin{tabular}{@{}lccc@{}}
\toprule
Metric & SixthSense & w/o CFM & CVAE \\
\midrule
DR $\uparrow$ (\%) & 86.70 & \textbf{91.04} & 79.95 \\
FPR$^{*}\downarrow$ (\%) & \textbf{0.66} & 8.05 & 6.47 \\
$\mathrm{LA}_{\mathrm{exact}}\uparrow$ (\%) & \textbf{66.89} & 33.10 & 51.57 \\
$R_{\mathrm{cov}}\uparrow$ (\%) & \textbf{57.14} & 29.96 & 40.89 \\
$e_F\downarrow$ (N) & \textbf{4.61} & 5.05 & 5.00 \\
$e_M\downarrow$ (N$\cdot$m) & \textbf{0.63} & 0.93 & 0.82 \\
\bottomrule
\end{tabular}
\endgroup

\caption{Real-robot contact-estimation results. Bold marks the best value
in each row.}
\label{tab:expbb}
\end{table}

\endgroup

\begingroup\color{blue}
\subsection{Multiple Plausible Contacts from One Observation}
\label{subsec:multimodal}
Beyond aggregate estimation metrics, we test whether repeated samples recover
distinct contact hypotheses under the same proprioceptive observation.

\textbf{Observation Ambiguity.}
Distinct contact locations and wrenches can produce indistinguishable
proprioceptive responses. For example, a force applied at the shared
joint of a parent--child link pair along the hinge axis produces the
same generalized effect whether assigned to the parent or child link
(Fig.~\ref{fig:ambiguity-construction}).
Generative models such as CVAE and SixthSense can represent multiple
dynamically plausible contact solutions for one observation, unlike
deterministic estimators such as w/o CFM and GMO. Unlike CPF, they do
not require dynamics inputs.

\begin{figure}[ht]
  \centering
  \includegraphics[width=1.0\columnwidth,trim={46bp 142bp 47bp 4bp},clip]{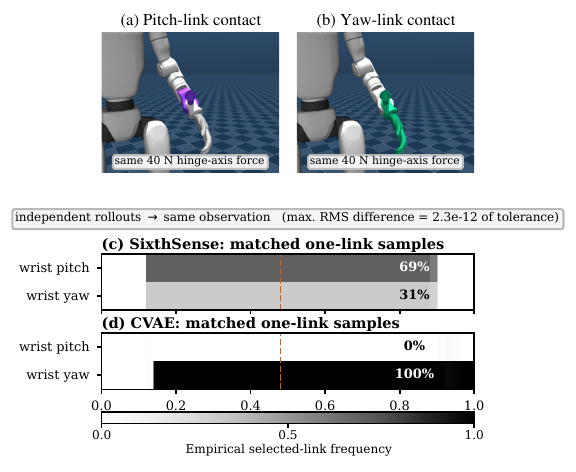}
  \caption{Equivalent contact
assignments produce the same observation.}
  \label{fig:ambiguity-construction}
\end{figure}

\begin{figure}[t]
  \centering
  \includegraphics[width=1.0\columnwidth]{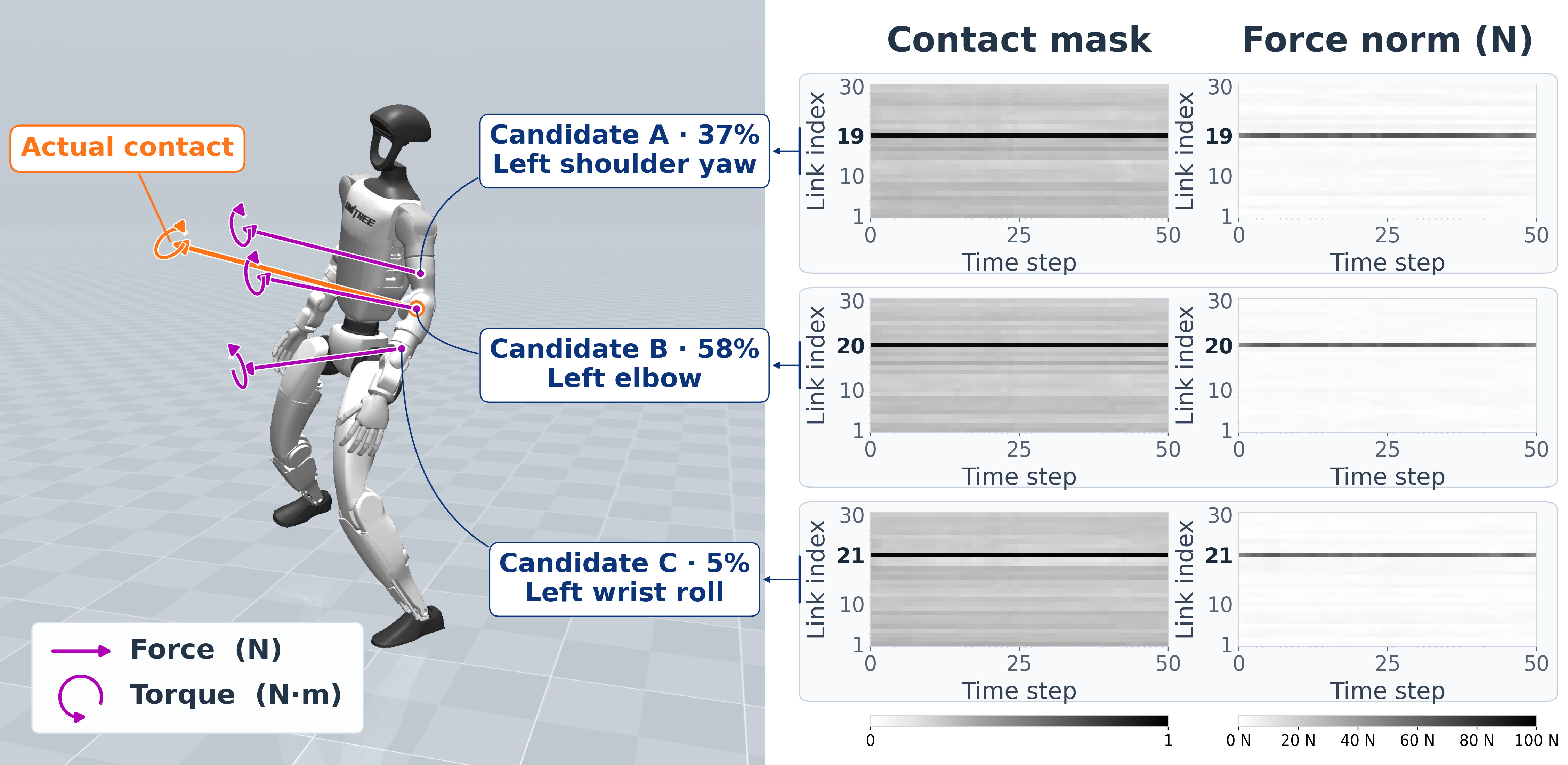}
  \caption{Multiple plausible contacts from one observation.
  Orange: ground truth; purple: sampled wrenches.
  Percentages show link frequencies among 200 samples.
  Heatmaps show raw gate scores and force norms before contact masking
  over time; gate-score shading does not represent probabilities.}
  \label{fig:ambiguity-example}
\end{figure}

\textbf{Multimodal Predictions from SixthSense.}
Fig.~\ref{fig:ambiguity-example} shows multiple plausible contact
estimates generated by SixthSense for a walking episode. Given the same
noisy observation, 200 samples assign contact to the left shoulder yaw,
left elbow, and left wrist roll with frequencies of 37\%, 58\%, and
5\%, respectively. The true
elbow contact receives the largest share. One sampled wrench per link
is visualized alongside its link activations and force norms.
These alternatives illustrate uncertainty in both contact location
and wrench: nearby contacts with different forces and moments can
produce similar proprioceptive responses under noise. By preserving
these alternatives, SixthSense represents uncertainty that a single
contact prediction cannot convey.

\textbf{Comparison with CVAE.}
We test 20 standing cases combining four parent--child hinge pairs and
five force magnitudes (30--50\,N). MuJoCo verifies two reference modes
per case with indistinguishable 50-frame observations. We draw 200 samples
from each method for each fixed observation, retaining CVAE's highest-gate
force-eligible link per frame. Stable modes require at least 5\% of
center-frame samples and presence in at least three of four sampling groups.
We report two-mode coverage ($R_{2\mathrm{mode}}$), the fraction of cases
with both modes stable; secondary-mode mass ($\bar m_{\mathrm{sec}}$),
the mean smaller reference-mode sample fraction across cases (zero if
either mode is absent); and local consistency ($R_{\mathrm{local}}$),
the fraction of assessable cases where at least half the samples from
each mode satisfy all six observation-channel replay tolerance bands
over the central 180\,ms. 

\begin{table}[h]
  \centering
  \caption{Empirical mode coverage and local consistency under controlled
  contact ambiguity. $R_{\mathrm{local}}$ uses the 18 cases with valid
  50\,Hz reference reconstruction.}
  \label{tab:ambiguity}
  \setlength{\tabcolsep}{4pt}
  \renewcommand{\arraystretch}{1.0}
  \small
  \begin{tabular}{lcc}
    \toprule
    \textbf{Statistic} & \textbf{CVAE} & \textbf{SixthSense} \\
    \midrule
    $R_{2\mathrm{mode}}$ & $4/20$ & $\mathbf{20/20}$ \\
    $\bar m_{\mathrm{sec}}$ & $1.7\%$ & $\mathbf{28.5\%}$ \\
    $R_{\mathrm{local}}$ & $3/18$ & $\mathbf{11/18}$ \\
    \bottomrule
  \end{tabular}
\end{table}

As Table~\ref{tab:ambiguity} shows, SixthSense achieves stable two-mode
coverage in 20/20 cases, versus 4/20 for CVAE, with higher secondary-mode
mass (28.5\% vs.\ 1.7\%) and local consistency (11/18 vs.\ 3/18).
Together, these results support better mode coverage and local replay
consistency in the tested cases, but do not measure posterior calibration.

\endgroup

\subsection{Applications: Physical Human--Robot Interaction}
\begin{figure}[ht]
    \centering
    \IfFileExists{figures/5_4_hri.png}{%
        \includegraphics[width=1.0\linewidth]{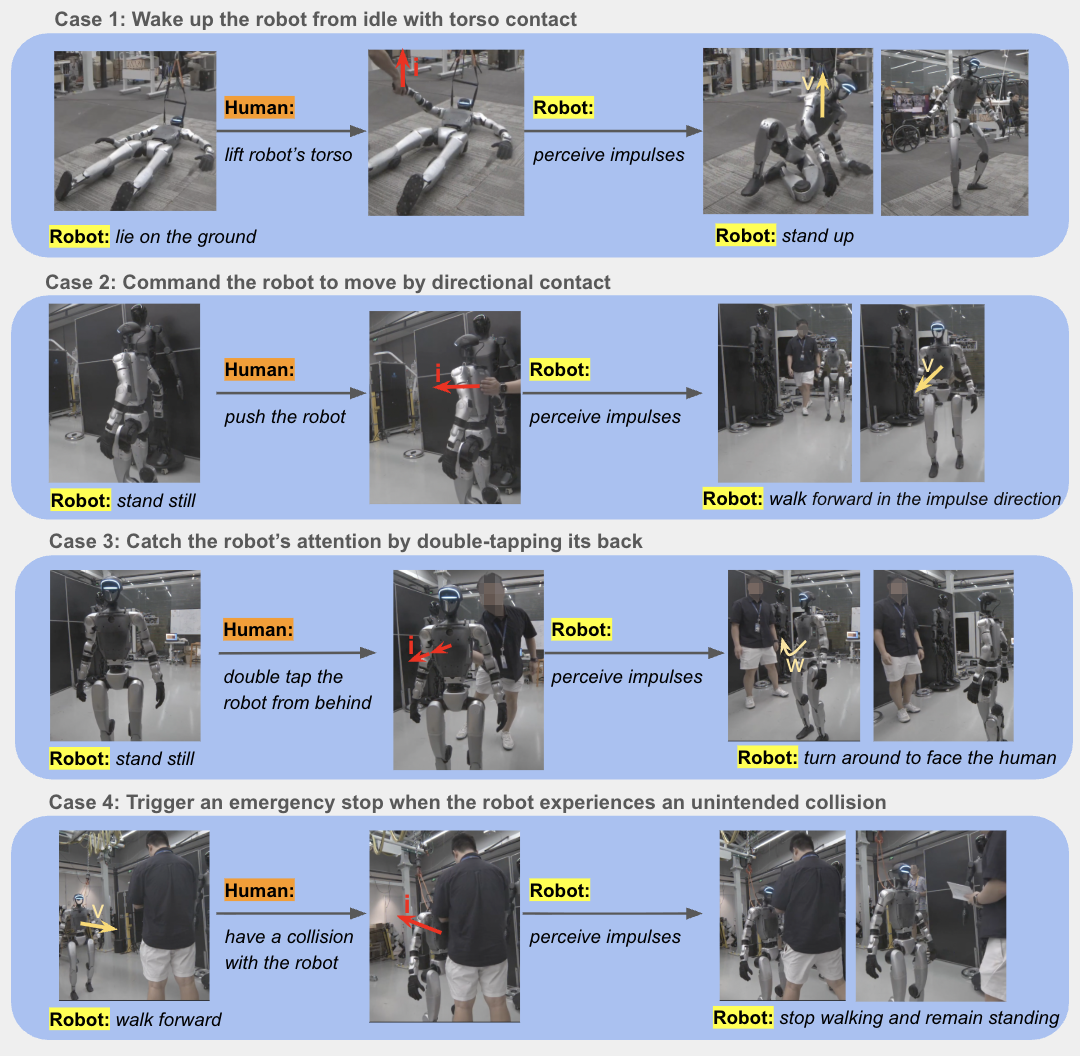}%
    }{%
        \fbox{\parbox[c][0.35\linewidth][c]{0.95\linewidth}{\centering Missing figure: figures/5_4_hri.png}}%
    }
    \caption{Physical human--robot interaction cases with SixthSense contact estimation}
    \label{fig:HRI}
\end{figure}

Whole-body contact perception is a key enabler for physical human--robot interaction (pHRI): inferring contacts from proprioception adds an interaction channel beyond vision, language, and teleoperation. \textcolor{blue}{Fig.~\ref{fig:HRI} illustrates three uses of this contact signal through simple state-machine demonstrations.}

\textbf{Safety.}
Improved whole-body contact perception enables earlier identification of hazardous situations, reducing both self-damage and risk to humans, as shown in Case~4.
This addresses the limited contact awareness of many humanoids over the torso and other non-manipulating regions.

\textbf{Contact as command.} When contact is reliably estimated and interpreted, it can serve as a direct and intuitive command channel.
As shown in Case~1, a user can pull the robot's arm to convey the intention to stand up, and the robot responds naturally without relying on speech or handheld controllers.

\textbf{Feedback for planning.} Our per-step output (a per-region contact mask with a wrench field) forms a structured, high-dimensional feedback signal that can be consumed by high-level planners to support environment-aware decision-making in pHRI.

\FloatBarrier
\section{Conclusion} 
\begingroup\color{blue}
We presented SixthSense, a shared estimator that infers humanoid whole-body
contact wrenches from proprioception using conditional flow matching.
We represent sparse, time-varying contacts as a spatiotemporal wrench field
and model ambiguous contact hypotheses as a distribution.
Simulation results demonstrate shared estimation across controller regimes
included in pooled training and recovery of multiple plausible contact
hypotheses under controlled ambiguity. On our collected real-robot dataset,
SixthSense outperforms learned baselines in contact estimation. Inference takes
approximately 50\,ms, and the estimator enables contact-triggered pHRI.

Limitations include reliance on proprioception alone and link-level spatial resolution.
Quantitative real-robot wrench evaluation is currently limited to standing, and future work will extend instrumented wrench validation to dynamic motions.
Shared weights across the evaluated regimes do not establish task-agnostic transfer to unseen controllers.
Single-contact training does not establish reliable zero-shot recovery of multi-contact sets.
Future work will test held-out controllers and motions, determine retraining needs, compare mixed-cardinality training, gather denser real-robot wrench data, and explore multimodal sensing and finer spatial estimates.
\par\endgroup













\bibliographystyle{IEEEtran}
\bibliography{references}

\end{document}